\documentclass{article}

\usepackage{spconf,amsmath,graphicx}

\usepackage{booktabs}
\usepackage{nicefrac}
\usepackage{microtype}      

\usepackage[T1]{fontenc}
\usepackage{amsmath,amssymb,amsfonts,mathrsfs,bm}
\usepackage{mathtools}
\usepackage{amsthm}
\usepackage{cite}
\usepackage[shortlabels]{enumitem}
\usepackage{graphicx}
\usepackage{epstopdf}
\usepackage{url}
\usepackage{colortbl}
\usepackage{multirow}
\usepackage[table,dvipsnames]{xcolor}
\usepackage[normalem]{ulem}
\usepackage{array}
\newcolumntype{L}[1]{>{\raggedright\let\newline\\\arraybackslash\hspace{0pt}}m{#1}}
\newcolumntype{C}[1]{>{\centering\let\newline\\\arraybackslash\hspace{0pt}}m{#1}}
\newcolumntype{R}[1]{>{\raggedleft\let\newline\\\arraybackslash\hspace{0pt}}m{#1}}

\makeatletter
\let\MYcaption\@makecaption
\makeatother
\usepackage[font=footnotesize]{subcaption}
\makeatletter
\let\@makecaption\MYcaption
\makeatother

\usepackage{xparse}



\usepackage{glossaries}

\makeatletter
\let\oldgls\gls
\let\oldglspl\glspl

\newcommand\fussy@ifnextchar[3]{%
  \let\reserved@d=#1%
  \def\reserved@a{#2}%
  \def\reserved@b{#3}%
  \futurelet\@let@token\fussy@ifnch}
\def\fussy@ifnch{%
  \ifx\@let@token\reserved@d
    \let\reserved@c\reserved@a 
  \else
    \let\reserved@c\reserved@b
  \fi
 \reserved@c}

\renewcommand{\gls}[1]{%
  \oldgls{#1}\fussy@ifnextchar.{\@checkperiod}{\@}}
\renewcommand{\glspl}[1]{%
  \oldglspl{#1}\fussy@ifnextchar.{\@checkperiod}{\@}}

\newcommand{\@checkperiod}[1]{%
  \ifnum\sfcode`\.=\spacefactor\else#1\fi
}
\makeatother

\newacronym{wrt}{w.r.t.}{with respect to}
\newacronym{RHS}{RHS}{right-hand side}
\newacronym{LHS}{LHS}{left-hand side}
\newacronym{iid}{i.i.d.}{independent and identically distributed}

\usepackage{float}

\ifx\notloadhyperref\undefined
	\ifx\loadbibentry\undefined
		\usepackage[hidelinks,hypertexnames=false]{hyperref} 
	\else
		\usepackage{bibentry}
		\makeatletter\let\saved@bibitem\@bibitem\makeatother
		\usepackage[hidelinks,hypertexnames=false]{hyperref}
		\makeatletter\let\@bibitem\saved@bibitem\makeatother
	\fi
\else
	\relax
\fi

\usepackage[capitalize]{cleveref}
\crefname{equation}{}{}
\Crefname{equation}{}{}
\crefname{claim}{claim}{claims}
\crefname{step}{step}{steps}
\crefname{line}{line}{lines}
\crefname{condition}{condition}{conditions}
\crefname{dmath}{}{}
\crefname{dseries}{}{}
\crefname{dgroup}{}{}

\crefname{Problem}{Problem}{Problems}
\crefformat{Problem}{Problem~(#2#1#3)}
\crefrangeformat{Problem}{Problems~(#3#1#4) to~(#5#2#6)}

\crefname{Theorem}{Theorem}{Theorems}
\crefname{Corollary}{Corollary}{Corollaries}
\crefname{Proposition}{Proposition}{Propositions}
\crefname{Lemma}{Lemma}{Lemmas}
\crefname{Definition}{Definition}{Definitions}
\crefname{Example}{Example}{Examples}
\crefname{Assumption}{Assumption}{Assumptions}
\crefname{Remark}{Remark}{Remarks}
\crefname{Rem}{Remark}{Remarks}
\crefname{remarks}{Remarks}{Remarks}
\crefname{Appendix}{Appendix}{Appendices}
\crefname{Exercise}{Exercise}{Exercises}
\crefname{Theorem_A}{Theorem}{Theorems}
\crefname{Corollary_A}{Corollary}{Corollaries}
\crefname{Proposition_A}{Proposition}{Propositions}
\crefname{Lemma_A}{Lemma}{Lemmas}
\crefname{Definition_A}{Definition}{Definitions}

\usepackage{algorithm,algorithmic}

\ifx\loadbreqn\undefined
	\relax
\else
	\usepackage{breqn} 
\fi

\ifx\renewtheorem\undefined
\ifx\useTheoremCounter\undefined
\newtheorem{Theorem}{Theorem}
\newtheorem{Corollary}{Corollary}
\newtheorem{Proposition}{Proposition}

\else

\fi

\fi

\theoremstyle{remark}

\theoremstyle{plain}

\newcommand{\ba}{\mathbf{a}}

\newcommand{\bW}{\mathbf{W}}

\DeclareSymbolFont{bsfletters}{OT1}{cmss}{bx}{n}
\DeclareSymbolFont{ssfletters}{OT1}{cmss}{m}{n}
\DeclareMathSymbol{\bsfGamma}{0}{bsfletters}{'000}
\DeclareMathSymbol{\ssfGamma}{0}{ssfletters}{'000}
\DeclareMathSymbol{\bsfDelta}{0}{bsfletters}{'001}
\DeclareMathSymbol{\ssfDelta}{0}{ssfletters}{'001}
\DeclareMathSymbol{\bsfTheta}{0}{bsfletters}{'002}
\DeclareMathSymbol{\ssfTheta}{0}{ssfletters}{'002}
\DeclareMathSymbol{\bsfLambda}{0}{bsfletters}{'003}
\DeclareMathSymbol{\ssfLambda}{0}{ssfletters}{'003}
\DeclareMathSymbol{\bsfXi}{0}{bsfletters}{'004}
\DeclareMathSymbol{\ssfXi}{0}{ssfletters}{'004}
\DeclareMathSymbol{\bsfPi}{0}{bsfletters}{'005}
\DeclareMathSymbol{\ssfPi}{0}{ssfletters}{'005}
\DeclareMathSymbol{\bsfSigma}{0}{bsfletters}{'006}
\DeclareMathSymbol{\ssfSigma}{0}{ssfletters}{'006}
\DeclareMathSymbol{\bsfUpsilon}{0}{bsfletters}{'007}
\DeclareMathSymbol{\ssfUpsilon}{0}{ssfletters}{'007}
\DeclareMathSymbol{\bsfPhi}{0}{bsfletters}{'010}
\DeclareMathSymbol{\ssfPhi}{0}{ssfletters}{'010}
\DeclareMathSymbol{\bsfPsi}{0}{bsfletters}{'011}
\DeclareMathSymbol{\ssfPsi}{0}{ssfletters}{'011}
\DeclareMathSymbol{\bsfOmega}{0}{bsfletters}{'012}
\DeclareMathSymbol{\ssfOmega}{0}{ssfletters}{'012}

\DeclarePairedDelimiter\parens{(}{)}
\DeclarePairedDelimiter\brk{[}{]}
\DeclarePairedDelimiter\braces{\{}{\}}

\newcommand{\qednew}{\nobreak \ifvmode \relax \else
      \ifdim\lastskip<1.5em \hskip-\lastskip
      \hskip1.5em plus0em minus0.5em \fi \nobreak
      \vrule height0.75em width0.5em depth0.25em\fi}

\newcommand{\T}{^{\intercal}}

\newcommand{\ofrac}[1]{{\frac{1}{#1}}}

\DeclareDocumentCommand\set{s m t| m}{%
  \IfBooleanTF#1%
	{\left\{\, #2\mathrel{} \IfBooleanTF{#3}{\middle|}{:}\mathrel{}  #4\, \right\}}%
  {\{\, #2 \IfBooleanTF{#3}{\mid}{\mathrel{} : \mathrel{}} #4\, \}}%
}

\DeclareDocumentCommand \ifcond {m m} {%
	{#1} %
	\IfValueT{#2}{\, \middle|\, {#2}}%
}

\DeclareDocumentCommand \P {e{_} g >{\SplitArgument{ 1 }{ @| }}d() g } {%
	\mathbb{P}%
	\IfValueTF{#1}{_{#1}}
		{\IfValueT{#2}{_{#2}}}%
	\IfValueT{#3}{\left(\ifcond#3}%
	\IfValueT{#4}{\, \middle|\, {#4}}%
	\IfValueT{#3}{\right)}%
}

\DeclareDocumentCommand \E {e{_} g >{\SplitArgument{ 1 }{ @| }}o g } {%
	\mathbb{E}%
	\IfValueTF{#1}{_{#1}}
		{\IfValueT{#2}{_{#2}}}%
	\IfValueT{#3}{\left[\ifcond#3}%
	\IfValueT{#4}{\, \middle|\, {#4}}%
	\IfValueT{#3}{\right]}%
}

\definecolor{gray90}{gray}{0.9}

\ifx\nohighlights\undefined

	\newcommand{\msout}[1]{\text{\color{green} \sout{\ensuremath{#1}}}}
	\newcommand{\del}[1]{{\color{green}\ifmmode \msout{#1}\else\sout{#1}\fi}}
\else

	\newcommand{\msout}[1]{#1}
	\newcommand{\del}[1]{#1}
\fi

\graphicspath{{./Figures/}} 
\ifx\diagnoselabel\undefined
	\relax
\else
	\makeatletter
	 \def\@testdef #1#2#3{%
		 \def\reserved@a{#3}\expandafter \ifx \csname #1@#2\endcsname
		\reserved@a  \else
	 \typeout{^^Jlabel #2 changed:^^J%
	 \meaning\reserved@a^^J%
	 \expandafter\meaning\csname #1@#2\endcsname^^J}%
	 \@tempswatrue \fi}
	\makeatother
\fi

\title{Learning on Heterogeneous Graphs Using High-Order Relations}
\name{See Hian Lee \qquad Feng Ji \qquad Wee Peng Tay}
\address{School of Electrical and Electronic Engineering, Nanyang Technological University, Singapore}

\begin{document}

\twocolumn[
\begin{@twocolumnfalse}

\copyright Copyright 2021 IEEE. Published in ICASSP 2021 - 2021 IEEE International Conference on Acoustics, Speech and Signal Processing (ICASSP), scheduled for 6-11 June 2021 in Toronto, Ontario, Canada. Personal use of this material is permitted. However, permission to reprint/republish this material for advertising or promotional purposes or for creating new collective works for resale or redistribution to servers or lists, or to reuse any copyrighted component of this work in other works, must be obtained from the IEEE. Contact: Manager, Copyrights and Permissions / IEEE Service Center / 445 Hoes Lane / P.O. Box 1331 / Piscataway, NJ 08855-1331, USA. Telephone: + Intl. 908-562-3966.\\


\end{@twocolumnfalse}
]

\topmargin=0mm
%
\maketitle
\begin{abstract}
A heterogeneous graph consists of different vertices and edges types. Learning on heterogeneous graphs typically employs meta-paths to deal with the heterogeneity by reducing the graph to a homogeneous network, guide random walks or capture semantics. These methods are however sensitive to the choice of meta-paths, with suboptimal paths leading to poor performance. In this paper, we propose an approach for learning on heterogeneous graphs without using meta-paths. Specifically, we decompose a heterogeneous graph into different homogeneous relation-type graphs, which are then combined to create higher-order relation-type representations. These representations preserve the heterogeneity of edges and retain their edge directions while capturing the interaction of different vertex types multiple hops apart. This is then complemented with attention mechanisms to distinguish the importance of the relation-type based neighbors and the relation-types themselves. Experiments demonstrate that our model generally outperforms other state-of-the-art baselines in the vertex classification task on three commonly studied heterogeneous graph datasets.\end{abstract}
\section{Introduction}
\label{sec:intro}

Learning representations that encode graph structural information is increasingly complex as we seek to model real-world graph data. In many applications, graph data are heterogeneous in nature. These heterogeneous graphs are innately ingrained with rich semantics and are more complex than their homogeneous counterparts \cite{zhang2019heterogeneous,cao2017meta}. Thus, existing approaches \cite{kipf2016semi,velivckovic2017graph} developed for homogeneous graphs have to be adapted or re-formularized to cater to the disparity and to capture the semantics present in a heterogeneous graph.

A typical approach is to use meta-paths \cite{hong2019attention,shi2016survey,yun2019gtn}. A meta-path \cite{sun2011pathsim} is a predefined sequence of vertex types. It is utilized to transform heterogeneous graphs into homogeneous graphs \cite{hong2019attention,yun2019gtn}, to guide random walks \cite{dong2017metapath2vec} or to capture the rich semantics \cite{sun2011pathsim}. This can be seen in techniques such as Heterogeneous graph Attention Network (HAN) \cite{han2019}, metapath2vec \cite{dong2017metapath2vec} and MEIRec \cite{fan2019metapath}. Meta-paths are widely utilized as they express different semantics, which help to mine the semantics information present \cite{han2019}. 

For instance, in a bibliographic graph, a meta-path ``APVPA'' , which refers to the vertex sequence ``author-paper-venue-paper-author'', conveys the idea that the authors published papers in the same venue. Meanwhile, ``APA'' indicates co-authorship relationship between two authors. However, selecting the optimal meta-paths for a specific task or dataset is a non-trivial task. It requires specialized domain knowledge regarding the relationships between different vertex types and edge types in a heterogeneous graph as well as the learning task at hand \cite{yun2019gtn,hu2020heterogeneous}. Constructing meta-paths based on domain knowledge becomes inefficient for graphs with a large number of vertex and edge types as the number of possible meta-paths grows exponentially with the number of vertex and edge types \cite{cao2017meta}. 

In this paper, we propose an approach to learn on heterogeneous graphs that does not use meta-paths. Instead, it leverages on the heterogeneity and directions of edges to build high-order relations that span multiple hops. An edge type in the graph is a first order relation-type. The heterogeneous graph is first decomposed into subgraphs with homogeneous first order relation-types. The adjacency matrices of these subgraphs are then multiplied together to form (weighted) adjacency matrices that represent higher-order relation-types amongst the vertices. Learning, together with attention mechanisms, is then performed on the set of higher-order relation-types. We call our approach lea\textbf{R}ning h\textbf{E}terogeneous \textbf{G}r\textbf{A}phs wi\textbf{T}h \textbf{H}igh-ord\textbf{E}r \textbf{R}elations (REGATHER). The acronym REGATHER invokes the idea of decomposing a difficult problem into basic units that can be combined to facilitate learning.

\begin{figure}[!htb]
\centering
	\includegraphics[width=0.74\linewidth]{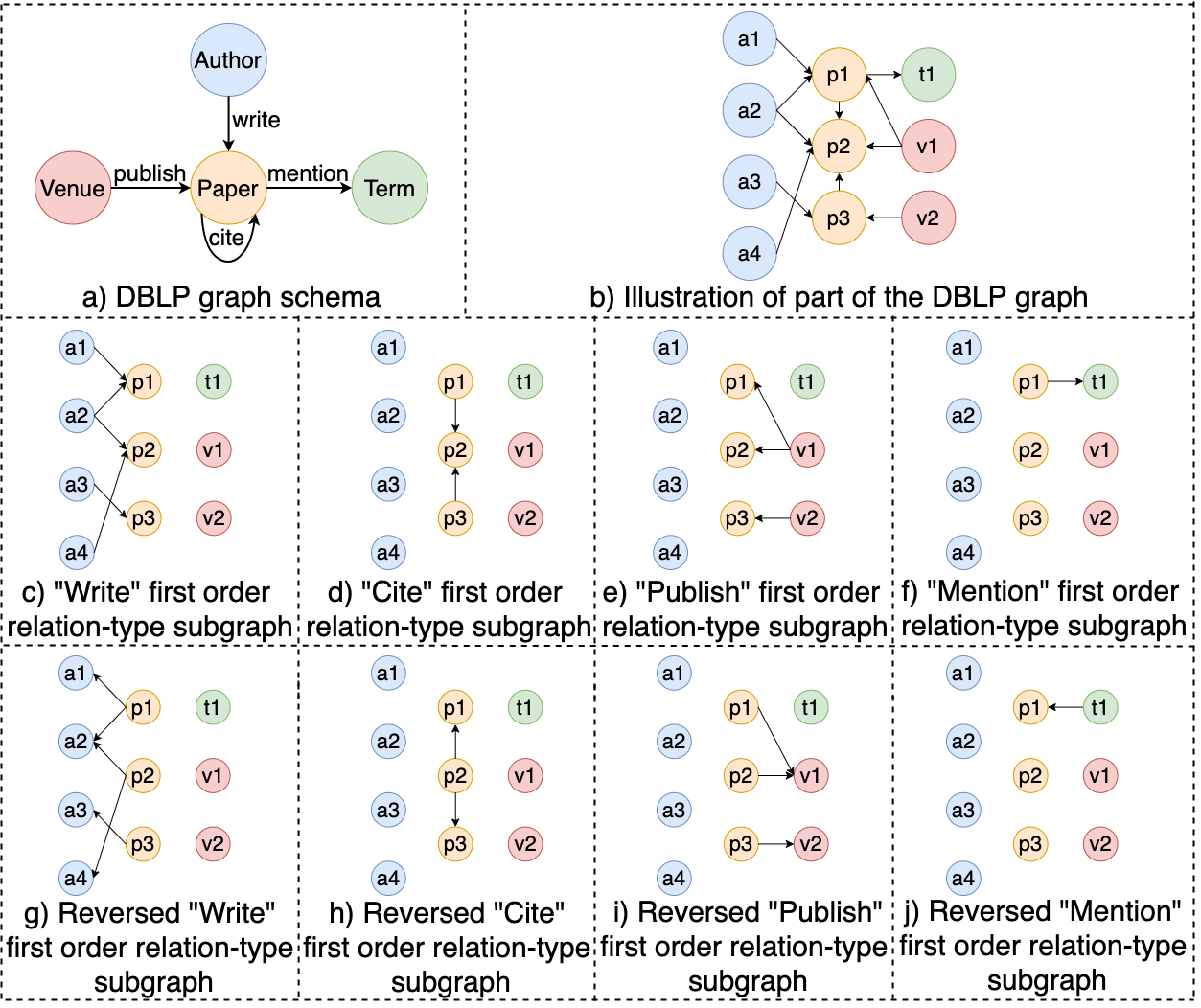}
\caption{Structure of DBLP graph: a) Schema, b) An illustration of DBLP graph, c)-f) First order relation-type subgraphs, g)-j) Reversed first order subgraphs. }\label{fig:DBLP}
\end{figure}

Taking DBLP dataset as example, the first order relation-type subgraphs are shown in \cref{fig:DBLP}c)-j). We consider the schema edge directions and their reverse directions as independent first order relation-types. In many existing studies \cite{kipf2016semi,velivckovic2017graph,chen2018fastgcn}, it is common to consider directed graphs as undirected \cite{gong2019exploiting,wu2020comprehensive}. However, this undesirably discards information captured in edge directions. By treating the reversed edge direction as an independent edge-type, we can learn different weights for an edge's forward and backward directions. 

\section{Related Work}
\label{sec:related work}

Learning on heterogeneous graphs with meta-paths have shown good results as seen in \cite{fan2019metapath, han2019,dong2017metapath2vec}. Yet, given the limitations of manually designing meta-paths, methods to automate the discovery of relevant meta-paths have been developed such as using greedy strategies \cite{meng2015discovering}, enumerating meta-paths within a fixed length \cite{lao2010relational}, using reinforcement learning \cite{yang2018similarity} and stacking multiple graph transformer layers \cite{yun2019gtn}. 

Learning on heterogeneous graphs based upon graph neural network (GNN) approaches that do not utilize meta-paths include Heterogeneous Graph Transformer (HGT) \cite{hu2020heterogeneous} and Heterogeneous Graph Neural Network (HetGNN) \cite{zhang2019heterogeneous}. These methods differ from our approach in their ways to incorporate information from multi-hop neighbors. HGT uses cross-layer information transfer to obtain information of neighbors that are multi-hop apart and HetGNN applies random walk with restart (RWR) strategy to sample all types of neighbor nodes from each node which implicitly learns higher-order relations. Moreover, HetGNN focuses on integrating heterogeneous content information of each node which differs from REGATHER that learns without incorporating content information.

Performing random walks on heterogeneous graphs without the use of meta-paths for guidance is also seen in \cite{just2018}. JUmp and STay (JUST) \cite{just2018} introduced a way to perform random walk on heterogeneous graphs without using meta-paths by using its jump and stay strategy to probabilistically balance \emph{jumping} to a different vertex type and \emph{staying} in the same vertex type during random walk. The whole process of JUST projects the vertices from different domains into a common vector space, generating vertex embeddings that are simply based upon structural information.

The attention mechanism is also evident in its usage in learning of graphs as seen in \cite{velivckovic2017graph,han2019,yun2019gtn}. Heterogeneous graph Attention Network (HAN) proposed by \cite{han2019} is a GNN designed for heterogeneous graphs employing hierarchical attention. Nevertheless, it uses a set of predefined meta-paths to transform a heterogeneous graph into a set of meta-path based homogeneous graphs. REGATHER employs dual-level attention like HAN but differs from HAN in three main ways. Firstly, instead of transforming a heterogeneous graph into meta-path based graphs, it transforms the heterogeneous graph into a set of relation-type subgraphs. Secondly, the heterogeneity of edges and their directions are retained, viewing the reverse edge direction as an independent first order relation-type. This is because even if two vertices are only connected via one direction semantically, they are related in the reverse direction from an information theoretic viewpoint. Lastly, in order to encode information from higher-order relations, which are again typically performed using customized meta-paths in the literature \cite{han2019}, REGATHER uses higher-order relations, which are discussed in \cref{subsec:relation-type}.

\section{Methodology}

In this section, we introduce the notion of relation-type subgraphs and present REGATHER's architecture. Throughout, we consider a heterogeneous directed graph $G = (V,E)$ where $V$ is the vertex set consisting of multiple vertices types, and $E$ is the set of edges consisting of multiple directed edge types.

\subsection{Relation-type subgraphs}\label{subsec:relation-type}

Each edge-type is defined to be a first order relation. We decompose the graph $G$ into a set of subgraphs $H_{1},\ldots, H_{c}$, where $c$ is the total number of edge-types and each subgraph $H_i$ has $|V|$ vertices and only one edge-type such that 
\begin{align*}
\bigcup_{1\leq i \leq c} H_{i} = G.
\end{align*}
For each $H_{i}$, there is an associated adjacency or relation matrix $M_{i}$. As the graph $G$ is directed, each matrix $M_i$ may not be symmetric. To consider edges with reversed directions, we take the transpose of each of the relation matrices of $H_{1}, \ldots, H_{c}$. Let $S$ be the set of matrices $\{M_{i} : 1\leq i \leq c\}$ and $S\T$ be the set of transposed matrices $\{M_{i}\T : 1\leq i \leq c\}$. We take their union as the first order relation matrices on $G$:
\begin{align*}
S^{1} = S\cup S\T.
\end{align*}
Each first order relation-type captures only partial information contained in the heterogeneous graph. For example, consider the DBLP graph in \cref{fig:DBLP}. There are no edges of the form ``author-author''. The fact that two authors may have co-authored a paper together, or attended the same conference previously, may contribute to our learning task. To capture such relationships, we consider higher-order relations by combining, e.g., ``author-paper'' and ``paper-author'' relation-type graphs. This is achieved by multiplication of the respective subgraph relation matrices together to obtain a (weighted) relation matrix. Nevertheless, it is possible to have some trivial resulting matrices given that some consecutive relation-types are unrealizable traversals. The trivial matrices are removed from the set. The $k$-th order relation-type is defined as
\begin{align*}
S^k = \braces*{ N_1N_2\cdots N_k : N_i \in S^1,\ 1\leq i \leq k}.
\end{align*}
We see that each meta-path of length $k$ corresponds to an edge in one of the relation matrices in $S^k$. Our goal is to perform learning over 
\begin{align*}
D = \bigcup_{k=1}^K S^k,
\end{align*}
for some $K\geq 1$. As an additional step, we add self-loops to each of the matrices in $D$ by adding an identity matrix $I$ to obtain the set of relation matrices
\begin{align}
\Phi =\braces*{A+I: A\in D}. 
\end{align}
Using high-order relation-type subgraphs, we eliminate the need to predefine meta-paths. Instead, these are exhaustively included in $\Phi$. In the following, we show how to use attention mechanisms to capture the importance of each relation-type in a data-driven fashion.

\subsection{Attention for relation-type based neighbors}\label{subsec:att-relation-type}

The relation matrices in $\Phi$ reflect the adjacency between vertices via the specific relation. Two vertices are considered adjacent if the entry in the matrix corresponding to the vertices is non-zero. The neighborhood of a vertex is the set of vertices adjacent to it. The different neighborhoods reflect the different aspects of a vertex. It is crucial to determine the importance of each of the adjacent vertices in a specific neighborhood because different adjacent vertices in a neighborhood can contribute differently to a vertex's representation. Hence, a single headed attention is applied to each of the relation matrices to determine the importance of a vertex's neighboring features specific to each relation matrix in $\Phi$. 

By using single headed graph attention, the number of parameters that have to be learned is reduced significantly. It is observed that reducing the attention heads to one did not negatively impact the performance of REGATHER. In some cases, it even improves REGATHER's performances. This is aligned with observations that multi-headedness plays an important role when dealing with specialized heads but most of the time, can be pruned for self-attention heads \cite{voita2019analyzing,michel2019sixteen}.

The input to each of the single headed graph attention layer is an relation matrix $\phi \in \Phi$ and a set of vertex features, $\mathbf{h} = \braces{h_{1}, h_{2},\ldots, h_{|V|}}$ where $h_i$ is the feature vector of vertex $i$ for $i=1,\ldots,|V|$. Given a vertex pair $(i,j)$, the unnormalized attention score between them can be learned as follows:

\begin{align}
e_{ij}^{\phi} = \mathrm{LeakyReLU}\parens*{\ba_{\phi}\T\brk*{\bW_{\phi}h_{i} \parallel \bW_{\phi} h_{j}}},
\end{align}
where $\parallel$ is the concatenation operation, $\ba_{\phi}$ represents the learnable attention vector for the respective $\phi$ and $\bW_{\phi}$ is the weight matrix for a linear transformation specific to the relation-type matrix $\phi$ and is shared by all the vertices. The attention vector $\ba_{\phi}$ depends on the input features, $\bW_{\phi}$ as well as the relation matrix $\phi$ that are input into the layer of single headed graph attention. The attention scores are then normalized by applying the softmax function to obtain the weight coefficients
\begin{align}
\alpha_{ij}^{\phi} = \mathrm{softmax}_{j}(e_{ij}^{\phi}) = \frac{\exp(e_{ij}^{\phi})}{\sum_{k\in N_{i}^{\phi}}\exp(e_{ik}^{\phi})},
\end{align}
where $N_{i}^{\phi}$ is the neighborhood of vertex $i$ in the specific relation-type matrix $\phi$. After the normalized weight coefficients are obtained, we can then employ these coefficients along with the corresponding neighbors' features to generate the output features. These output features are specific to $\phi$ for each of the vertices $i=1,\ldots,|V|$, and given as
\begin{align}
z_{i}^{\phi} = \sigma\parens*{\sum_{j\in N_{i}^{\phi}} \alpha_{ij}^{\phi} \mathbf{W}_{\phi}h_{j}},
\end{align}
where $\sigma$ denotes an activation function. For each $\phi\in\Phi$, the output is a set of relation-type specific vertex embeddings, $Z_{\phi} = \braces*{z_{1}^\phi,\ldots,z_{|V|}^\phi}$.

\subsection{Attention to fuse different relation-types}

The new vertex embeddings $\braces{Z_{\phi} : \phi\in\Phi}$ reflect the different aspects of the vertices under different relation-types $\phi$ and have to be fused. A second attention layer accounts for the different contribution of each relation-type in describing the vertices, giving more weight to the more crucial relation-type and less weights to the less salient relation-type. 

\begin{figure}[!htb]
\centering
	\includegraphics[width=1\linewidth]{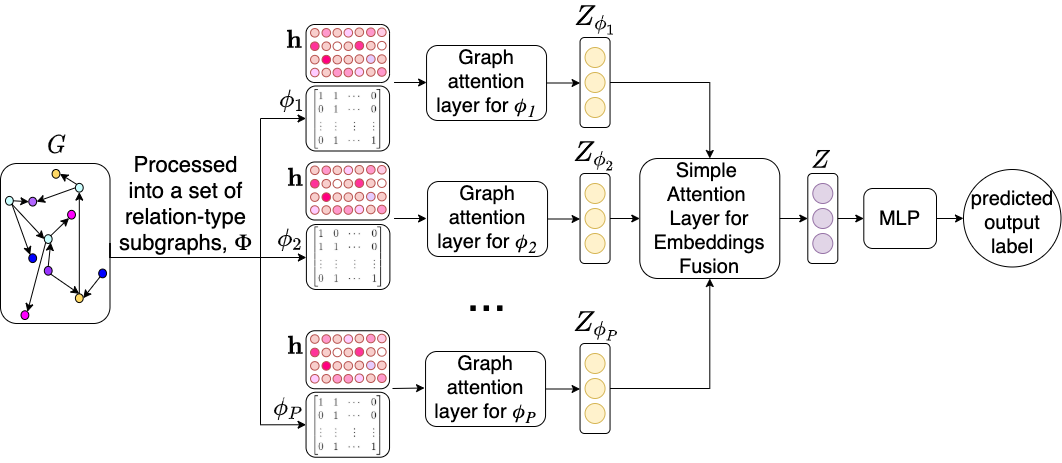}
\caption{Overall architecture of REGATHER.
}\label{fig:REGATHER}
\end{figure}

For each $\phi\in\Phi$, the relation-type specific vertex embeddings are transformed via a linear transformation consisting of a weight matrix $\mathbf{F}$ and a bias vector $\mathbf{b}$, then followed by a $\tanh$ function. The unnormalized attention score for each of the relation-type specific embeddings is then learnt as follows:
\begin{align}
w_{\phi} = \ofrac{|V|}\sum_{i \in V} \mathbf{q}\T \tanh(\mathbf{F}z_{i}^{\phi}+\mathbf{b}),
\end{align}
where $\mathbf{q}$ refers to the learnable relation-type specific attention vector. The second attention layer employs averaging instead of concatenation which means that the importance of each of the relation-type $w_{\phi}$ where $\phi\in\Phi$, is generated as a weighted average of all the vertex embeddings within a group of relation-type specific embeddings. After $w_{\phi}$ is obtained, the softmax function is then applied. The normalized values are the weights of the respective relation-type denoted as $\beta_{\phi}$:
\begin{align}
\beta_{\phi} = \frac{\exp(w_{\phi})}{\sum_{\phi'\in\Phi} \exp(w_{\phi'})}.
\end{align}
The different relation-type specific vertex embeddings are then fused to generate the final vertex embedding via linear combination using the learned weights as follows:
\begin{align}\label{eq:Z}
Z = \sum_{\phi\in\Phi} \beta_{\phi} Z_{\phi}.
\end{align}
The vertex embedding $Z$ can then be applied to various tasks, optimising the model via backpropagation. Suppose that $|\Phi| = P$, the architecture of REGATHER is as shown in \cref{fig:REGATHER}.

\section{Experiments}

We perform experiments on three heterogeneous graph datasets, similar to those utilized in \cite{just2018} and evaluate REGATHER on vertex classification tasks. Our objective is to predict the labels of a target vertex type. The statistics of the datasets utilised are as shown in \cref{table:stats}. For the loss function, cross-entropy loss is employed. Only the vertices belonging to the target type is considered when measuring cross entropy $L$ 
\begin{align*}
L = - \sum_{l\in y_{L}} Y(l)\ln(\sigma{(Z(l))}),
\end{align*}
where $\sigma$ denotes an activation function, $y_{L}$ is the set of target vertex type's indices with labels, $Y(l)$ and $Z(l)$ are the labels and learned embeddings of the labeled vertex $l$, respectively. 

\begin{table}[!htb]
\caption{Statistics of the heterogeneous graph datasets utilized.}\label{table:stats}
\centering
\resizebox{\columnwidth}{!}{
\begin{tabular}{llcccccc}
\hline
Dataset                     & Relations (A-B)               & Feature              & \#Label            & \#Vertex                 & \#Edge                 & \#Heterogeneous Edge   & \#Homogeneous Edge     \\ \hline
\multirow{4}{*}{DBLP}       & Paper-Paper                   & \multirow{4}{*}{334} & \multirow{4}{*}{4}  & \multirow{4}{*}{15,649} & \multirow{4}{*}{51,377} & \multirow{4}{*}{44,379} & \multirow{4}{*}{6,998}  \\
                            & Author-Paper                  &                      &                     &                         &                         &                         &                         \\
                            & Venue-Paper                   &                      &                     &                         &                         &                         &                         \\
                            & Paper-Term                    &                      &                     &                         &                         &                         &                         \\ \hline
\multirow{5}{*}{Movie}      & Movie-Movie                   & \multirow{5}{*}{334} & \multirow{5}{*}{5}  & \multirow{5}{*}{21,345} & \multirow{5}{*}{89,038} & \multirow{5}{*}{34,354} & \multirow{5}{*}{54,684} \\
                            & Actor-Actor                   &                      &                     &                         &                         &                         &                         \\
                            & Actor-Movie                   &                      &                     &                         &                         &                         &                         \\
                            & Director-Movie                &                      &                     &                         &                         &                         &                         \\
                            & Composer-Movie                &                      &                     &                         &                         &                         &                         \\ \hline
\multirow{4}{*}{Foursquare} & User-User                     & \multirow{4}{*}{128} & \multirow{4}{*}{10} & \multirow{4}{*}{29,771} & \multirow{4}{*}{83,407} & \multirow{4}{*}{77,712} & \multirow{4}{*}{5,695}  \\
                            & User-Check\_in                &                      &                     &                         &                         &                         &                         \\
                            & Check\_in-Time\_slot          &                      &                     &                         &                         &                         &                         \\
                            & Check\_in-Point\_of\_interest &                      &                     &                         &                         &                         &                        
\\ \bottomrule
\end{tabular}}
\end{table}

We compare our model with JUST \cite{just2018}, GAT \cite{velivckovic2017graph} and HAN \cite{han2019}, which are state-of-the-art methods to demonstrate the effectiveness of our proposed model. For GAT, the heterogeneity in vertex types and edge types are ignored. JUST is evaluated using the same settings as in \cite{just2018}. For the observed vertex feature vectors, we use the JUST pre-trained embeddings for all methods to enable fair comparison as this ensures that each method's performance is not due to engineered features that might have incorporated additional information. Nevertheless, we note that engineered features like those used in \cite{han2019} can also be used with REGATHER.

The input to JUST is the heterogeneous graph and it outputs embeddings that are used as the initial features for all the other methods under comparison. The parameters in JUST like the embeddings' dimension $d$ , the SkipGram model's window size $k$, the memorized domains $m$ and the initial stay parameter $\gamma$, are optimized for the different datasets. For REGATHER, GAT and HAN, we evaluated all of them using the same setting as in \cite{han2019}. The differences are that REGATHER uses single head attention and the highest order of relation-type $K$ in REGATHER is set to 3. In brief, we trained the models for a maximum of 200 epochs using Adam with a learning rate of 0.005 and early stopping with a window size of 100. The relation-type specific attention vector $\mathbf{q}$ in HAN and REGATHER is set to a dimension of 128. 

The experimental results are summarised in \cref{table:results}. We report the average Macro-F1 and Micro-F1 scores (with standard deviation) from 10 repeated trials. Our results demonstrate that our method considerably improves JUST pre-trained embeddings and in most cases, yield better results in vertex classification tasks compared to other baselines.

\begin{table}[!htb]
\caption{Quantitative results (\%) of vertex classification task.}\label{table:results}
\centering
\resizebox{\columnwidth}{!}{
\begin{tabular}{@{}llccccc@{}}
\toprule
Datasets                    & Metrics                   & Training Size & JUST                 & GAT               & HAN               & REGATHER \\ \midrule
\multirow{8}{*}{DBLP}       & \multirow{4}{*}{Macro-F1} & 20\%          & 84.41 $\pm$ 0.68     & 88.25$\pm$1.15    & 87.15$\pm$0.90    & {\bf 89.04$\pm$0.51}         \\
                            &                           & 40\%          & 84.52 $\pm$ 0.90     & 89.22$\pm$1.03    & 88.51$\pm$1.72    & {\bf 90.87$\pm$0.95}         \\
                            &                           & 60\%          & 85.06 $\pm$ 1.07     & 89.40$\pm$1.70    & 89.21$\pm$2.02    & {\bf 92.25$\pm$2.00}         \\
                            &                           & 80\%          & 85.28 $\pm$ 1.63     & 89.54$\pm$2.61    & 90.13$\pm$1.26    & {\bf 92.66$\pm$1.70}         \\ \cmidrule(l){2-7}
                            & \multirow{4}{*}{Micro-F1} & 20\%          & 84.41 $\pm$ 0.67     & 88.23$\pm$1.15    & 87.15$\pm$0.89    & {\bf 89.04$\pm$0.52}         \\
                            &                           & 40\%          & 84.53 $\pm$ 0.90     & 89.20$\pm$1.05    & 88.53$\pm$1.72    & {\bf 90.85$\pm$0.97}         \\
                            &                           & 60\%          & 85.09 $\pm$ 1.07     & 89.41$\pm$1.71    & 89.25$\pm$2.04    & {\bf 92.25$\pm$2.01}         \\
                            &                           & 80\%          & 85.29 $\pm$ 1.67     & 89.53$\pm$2.61    & 90.16$\pm$1.35    & {\bf 92.62$\pm$1.73}         \\
\midrule
\multirow{8}{*}{Movie}      & \multirow{4}{*}{Macro-F1} & 20\%          & 35.88 $\pm$ 0.86     & 40.64$\pm$1.29    & 39.36$\pm$2.82    & {\bf 46.87$\pm$4.04}         \\
                            &                           & 40\%          & 39.12 $\pm$ 0.86     & 45.87$\pm$2.07    & 49.62$\pm$2.43    & {\bf 54.68$\pm$3.03}         \\
                            &                           & 60\%          & 40.18 $\pm$ 0.64     & 47.07$\pm$4.09    & 53.22$\pm$2.71    & {\bf 57.94$\pm$2.72}         \\
                            &                           & 80\%          & 41.07 $\pm$ 1.18     & 48.68$\pm$2.49    & 54.40$\pm$3.03    & {\bf 58.44$\pm$3.82}         \\ \cmidrule(l){2-7}
                            & \multirow{4}{*}{Micro-F1} & 20\%          & 38.19 $\pm$ 0.67     & 49.67$\pm$0.72    & 48.35$\pm$1.33    & {\bf 55.07$\pm$2.05}         \\
                            &                           & 40\%          & 42.74 $\pm$ 0.67     & 54.08$\pm$1.33    & 56.88$\pm$2.05    & {\bf 60.93$\pm$2.25}         \\
                            &                           & 60\%          & 44.11 $\pm$ 0.64     & 56.63$\pm$2.83    & 60.09$\pm$2.13    & {\bf 65.29$\pm$1.99}         \\
                            &                           & 80\%          & 45.11 $\pm$ 0.94     & 58.81$\pm$2.38    & 62.84$\pm$3.07    & {\bf 66.50$\pm$3.08}         \\
\midrule
\multirow{8}{*}{Foursquare} & \multirow{4}{*}{Macro-F1} & 20\%          & 32.53 $\pm$ 2.14     & 46.74$\pm$1.46    & {\bf 56.11$\pm$2.42}    & 49.47$\pm$2.88         \\
                            &                           & 40\%          & 38.09 $\pm$ 1.73     & 59.49$\pm$2.78    & {\bf 64.33$\pm$2.25}    & 61.13$\pm$3.79         \\
                            &                           & 60\%          & 41.65 $\pm$ 2.99     & 67.76$\pm$3.16    & 66.73$\pm$4.82    & {\bf 68.55$\pm$5.34}         \\
                            &                           & 80\%          & 46.49 $\pm$ 3.41     & 68.24$\pm$4.75    & 66.39$\pm$5.48    & {\bf 69.19$\pm$6.13}         \\ \cmidrule(l){2-7}
                            & \multirow{4}{*}{Micro-F1} & 20\%          & 40.17 $\pm$ 1.73     & 52.12$\pm$1.28    & {\bf 61.48$\pm$1.44}    & 56.29$\pm$1.56         \\ 
                            &                           & 40\%          & 44.13 $\pm$ 1.60     & 63.94$\pm$1.91    & {\bf 68.80$\pm$1.78}    & 66.45$\pm$2.46         \\
                            &                           & 60\%          & 47.08 $\pm$ 3.17     & 71.36$\pm$2.73    & 70.29$\pm$4.17    & {\bf 72.77$\pm$4.12}         \\
                            &                           & 80\%          & 51.76 $\pm$ 2.28     & 72.00$\pm$3.56    & 71.04$\pm$2.96    & {\bf 73.68$\pm$4.34}         \\ \bottomrule
\end{tabular}}
\end{table}



\section{Conclusion}

In this paper, we explored learning on heterogeneous graphs without using meta-paths as optimal meta-path selection is problematic especially when the dataset is complex and has many vertex types. Our method REGATHER uses relation-type subgraphs to learn representations and to capture interactions multiple hops apart. We also use dual-level attention to learn the importance of neighboring vertices and different relation-types. Experimental results demonstrated REGATHER's effectiveness in improving JUST pre-embeddings compared to GAT and HAN across three datasets.


\vfill\pagebreak

\bibliographystyle{IEEEbib}
\bibliography{refs}

\begin{thebibliography}{10}

\bibitem{zhang2019heterogeneous}
Chuxu Zhang, Dongjin Song, Chao Huang, Ananthram Swami, and Nitesh~V Chawla,
\newblock ``Heterogeneous graph neural network,''
\newblock in {\em Proceedings of the ACM SIGKDD International Conference on
  Knowledge Discovery and Data Mining}, 2019, pp. 793--803.

\bibitem{cao2017meta}
Xiaohuan Cao, Yuyan Zheng, Chuan Shi, Jingzhi Li, and Bin Wu,
\newblock ``Meta-path-based link prediction in schema-rich heterogeneous
  information network,''
\newblock {\em International Journal of Data Science and Analytics}, vol. 3,
  no. 4, pp. 285--296, 2017.

\bibitem{kipf2016semi}
Thomas~N Kipf and Max Welling,
\newblock ``Semi-supervised classification with graph convolutional networks,''
\newblock in {\em International Conference on Learning Representations}, 2016.

\bibitem{velivckovic2017graph}
Petar Veli{\v{c}}kovi{\'c}, Guillem Cucurull, Arantxa Casanova, Adriana Romero,
  Pietro Lio, and Yoshua Bengio,
\newblock ``Graph attention networks,''
\newblock in {\em International Conference on Learning Representations}, 2017.

\bibitem{hong2019attention}
Huiting Hong, Hantao Guo, Yucheng Lin, Xiaoqing Yang, Zang Li, and Jieping Ye,
\newblock ``An attention-based graph neural network for heterogeneous
  structural learning,''
\newblock {\em arXiv preprint arXiv:1912.10832}, 2019.

\bibitem{shi2016survey}
Chuan Shi, Yitong Li, Jiawei Zhang, Yizhou Sun, and S~Yu Philip,
\newblock ``A survey of heterogeneous information network analysis,''
\newblock {\em IEEE Transactions on Knowledge and Data Engineering}, vol. 29,
  no. 1, pp. 17--37, 2016.

\bibitem{yun2019gtn}
Seongjun Yun, Minbyul Jeong, Raehyun Kim, Jaewoo Kang, and Hyunwoo~J Kim,
\newblock ``Graph transformer networks,''
\newblock in {\em Advances in Neural Information Processing Systems}, 2019, pp.
  11960--11970.

\bibitem{sun2011pathsim}
Yizhou Sun, Jiawei Han, Xifeng Yan, Philip~S Yu, and Tianyi Wu,
\newblock ``Pathsim: {M}eta path-based top-k similarity search in heterogeneous
  information networks,''
\newblock in {\em Proceedings of the International Conference on Very Large
  Data Based}, 2011, pp. 992--1003.

\bibitem{dong2017metapath2vec}
Yuxiao Dong, Nitesh~V Chawla, and Ananthram Swami,
\newblock ``Metapath2vec: {S}calable representation learning for heterogeneous
  networks,''
\newblock in {\em Proceedings of the ACM SIGKDD International Conference on
  Knowledge Discovery and Data Mining}, 2017, pp. 135--144.

\bibitem{han2019}
Xiao Wang, Houye Ji, Chuan Shi, Bai Wang, Yanfang Ye, Peng Cui, and Philip~S
  Yu,
\newblock ``Heterogeneous graph attention network,''
\newblock in {\em Proceedings of the International World Wide Web Conference},
  2019, pp. 2022--2032.

\bibitem{fan2019metapath}
Shaohua Fan, Junxiong Zhu, Xiaotian Han, Chuan Shi, Linmei Hu, Biyu Ma, and
  Yongliang Li,
\newblock ``Metapath-guided heterogeneous graph neural network for intent
  recommendation,''
\newblock in {\em Proceedings of the ACM SIGKDD International Conference on
  Knowledge Discovery and Data Mining}, 2019, pp. 2478--2486.

\bibitem{hu2020heterogeneous}
Ziniu Hu, Yuxiao Dong, Kuansan Wang, and Yizhou Sun,
\newblock ``Heterogeneous graph transformer,''
\newblock in {\em Proceedings of the International World Wide Web Conference},
  2020, pp. 2704--2710.

\bibitem{chen2018fastgcn}
Jie Chen, Tengfei Ma, and Cao Xiao,
\newblock ``Fast{GCN}: {F}ast learning with graph convolutional networks via
  importance sampling,''
\newblock {\em arXiv preprint arXiv:1801.10247}, 2018.

\bibitem{gong2019exploiting}
Liyu Gong and Qiang Cheng,
\newblock ``Exploiting edge features for graph neural networks,''
\newblock in {\em Proceedings of the IEEE Conference on Computer Vision and
  Pattern Recognition}, 2019, pp. 9211--9219.

\bibitem{wu2020comprehensive}
Zonghan Wu, Shirui Pan, Fengwen Chen, Guodong Long, Chengqi Zhang, and S~Yu
  Philip,
\newblock ``A comprehensive survey on graph neural networks,''
\newblock {\em IEEE Transactions on Neural Networks and Learning Systems},
  2020.

\bibitem{meng2015discovering}
Changping Meng, Reynold Cheng, Silviu Maniu, Pierre Senellart, and Wangda
  Zhang,
\newblock ``Discovering meta-paths in large heterogeneous information
  networks,''
\newblock in {\em Proceedings of the International Conference on World Wide
  Web}, 2015, pp. 754--764.

\bibitem{lao2010relational}
Ni~Lao and William~W Cohen,
\newblock ``Relational retrieval using a combination of path-constrained random
  walks,''
\newblock {\em Machine Learning}, pp. 53--67, 2010.

\bibitem{yang2018similarity}
Carl Yang, Mengxiong Liu, Frank He, Xikun Zhang, Jian Peng, and Jiawei Han,
\newblock ``Similarity modeling on heterogeneous networks via automatic path
  discovery,''
\newblock in {\em Joint European Conference on Machine Learning and Knowledge
  Discovery in Databases}. Springer, 2018, pp. 37--54.

\bibitem{just2018}
Rana Hussein, Dingqi Yang, and Philippe Cudr{\'e}-Mauroux,
\newblock ``Are meta-paths necessary? {R}evisiting heterogeneous graph
  embeddings,''
\newblock in {\em Proceedings of the ACM International Conference on
  Information and Knowledge Management}, 2018, pp. 437--446.

\bibitem{voita2019analyzing}
Elena Voita, David Talbot, Fedor Moiseev, Rico Sennrich, and Ivan Titov,
\newblock ``Analyzing multi-head self-attention: {S}pecialized heads do the
  heavy lifting, the rest can be pruned,''
\newblock {\em arXiv preprint arXiv:1905.09418}, 2019.

\bibitem{michel2019sixteen}
Paul Michel, Omer Levy, and Graham Neubig,
\newblock ``Are sixteen heads really better than one?,''
\newblock in {\em Advances in Neural Information Processing Systems}, 2019, pp.
  14014--14024.

\end{thebibliography}
\end{document}